\documentclass{article}
\usepackage{spconf,amsmath,graphicx}
\usepackage{amsfonts}
\usepackage{algorithm, algorithmic, amsthm, paralist}
\usepackage{booktabs}

\usepackage{todonotes}

\title{Adversarial Advantage Actor-Critic Model for \\Task-Completion Dialogue Policy Learning}
\name{Baolin Peng$^{\star}$\quad Xiujun Li$^{\dagger}$\quad Jianfeng Gao$^{\dagger}$\quad Jingjing Liu$^{\dagger}$\quad Yun-Nung Chen$^{\ddag}$\quad Kam-Fai Wong$^{\star}$} 
\address{$^{\dagger}$Microsoft Research, Redmond, WA, USA \\
  $^{\star}$The Chinese University of Hong Kong, Hong Kong \\
  $^{\ddag}$National Taiwan University, Taipei, Taiwan
}

\begin{document}
%\ninept
%
\maketitle
\begin{abstract}
This paper presents a new method --- adversarial advantage actor-critic (Adversarial A2C), which significantly improves the efficiency of dialogue policy learning in task-completion dialogue systems. Inspired by generative adversarial networks (GAN), we train a discriminator to differentiate responses/actions generated by dialogue agents from responses/actions by experts. Then, we incorporate the discriminator as another critic into the advantage actor-critic (A2C) framework, to encourage the dialogue agent to explore state-action within the regions where the agent takes actions similar to those of the experts. Experimental results in a movie-ticket booking domain show that the proposed Adversarial A2C can accelerate policy exploration efficiently.

%When the reward function is known, the goal of reinforcement learning is to find a policy $\pi$ which can maximize the cumulative reward. While, in many of cases, the reward function is unknown, designing good reward functions is a challenging problem, and furthermore the choice of reward function will affect the policy learning. In the task-completion dialogue systems, conventional approaches for reward function are designed with domain knowledge, and also the handcrafted reward function does not leverage the experience gained online during the policy learning. In this paper, we propose an Adversarial Advantage Actor-Critic (Adversarial A2C) model, which can learn a parameterized reward function with online experience, and then learn a dialogue policy with reinforcement learning in the task-completion dialogue systems. Empirical results on a  movie-ticket booking domain show that the proposed model can improve the dialogue policy learning in terms of both learning speed and performance.

\end{abstract}
\section{Introduction}
\label{sec:intro}
There has been growing interest in exploiting reinforcement learning (RL) for policy learning in task-oriented dialogue systems ~\cite{young2013pomdp,fatemi2016policy,zhao2016towards,su2016continuously,li2017end,williams2017hybrid,dhingra2017towards,peng2017composite,liu2017iterative,peng2018integrating}. One of the biggest challenges in these approaches is the \emph{reward sparsity} issue. Dialogue policy learning for complex tasks, such as movie-ticket booking and travel planning, requires exploration in a large state-action space, and it often takes many conversation turns between the user and the agent to fulfill a task, leading to a long trajectory. Thus, the reward signals (usually provided by users at the end of a conversation) are often delayed and sparse.

To deal with reward sparsity, different approaches have been proposed recently, with promising empirical results. One approach is to leverage prior knowledge learned from expert-generated (or human-human) dialogue. For example, instead of learning a dialogue policy from scratch, we construct an initial policy learned from human-human dialogues, via imitation learning or hand-crafted rules. Prior work showed that a pre-trained supervised policy or a weak rule-based policy can significantly improve the efficiency of exploration~\cite{su2016continuously,lipton2016efficient}.
Another approach is to introduce heuristics, often in the form of the intrinsic reward to guide the exploration~\cite{chentanez2005intrinsically,mohamed2015variational,houthooft2016vime,jaderberg2016reinforcement}.
While the extrinsic reward (e.g., feedback provided by users at the end of a conversation) could be sparse, it is possible to get intrinsic reward after each action in order to guide the agent to explore the region more effectively. For example, VIME maximizes information gain about the agent’s belief of environment dynamics~\cite{houthooft2016vime}. It adds an intrinsic reward bonus to the reward function, which quantifies the agent’s \emph{surprise} to encourage the agent to explore the regions that are relatively unexplored.
BBQN encourages the agent to explore those state-action regions where the agent is relatively uncertain in action selection~\cite{lipton2016efficient}. UNREAL converts the training signals from three auxiliary tasks as intrinsic rewards, which significantly improved the learning speed and the robustness of the agent~\cite{jaderberg2016reinforcement}.

In this paper, we present a new method that combines the strength of the two approaches mentioned above. Similar to the first approach, we also leverage expert-generated dialogues as prior knowledge. However, instead of constructing an initial dialogue policy using prior knowledge, we, inspired by generative adversarial networks (GAN)~\cite{goodfellow2014generative}, train a discriminator to differentiate the responses (or actions) generated by dialogue agents from those by human experts. Then, we use the output of the discriminator as intrinsic reward to encourage the dialogue agent to explore state-action regions in which the agent takes actions similar to what human experts do. Specifically, we incorporate the discriminator as another critic into the advantage actor-critic (A2C) framework, resulting in a new model, called adversarial advantage actor-critic (Adversarial A2C). The modeling assumption behind our method is that the expert policies (embedded in the expert-generated dialogues) are reasonably good, thus the agent-selected actions, which are more similar to expert-selected ones, lead more often to successful dialogues with positive rewards. In a word, we remedy the reward sparse problem on two fronts, by leveraging human-human dialogues as prior knowledge and by introducing intrinsic rewards. Experiments in a movie-ticket booking domain show that the proposed Adversarial A2C model can significantly improve dialogue policy learning in terms of both effectiveness and efficiency.

\section{Methodology}

\begin{figure}[t]
\centering
\includegraphics[width=1\linewidth]{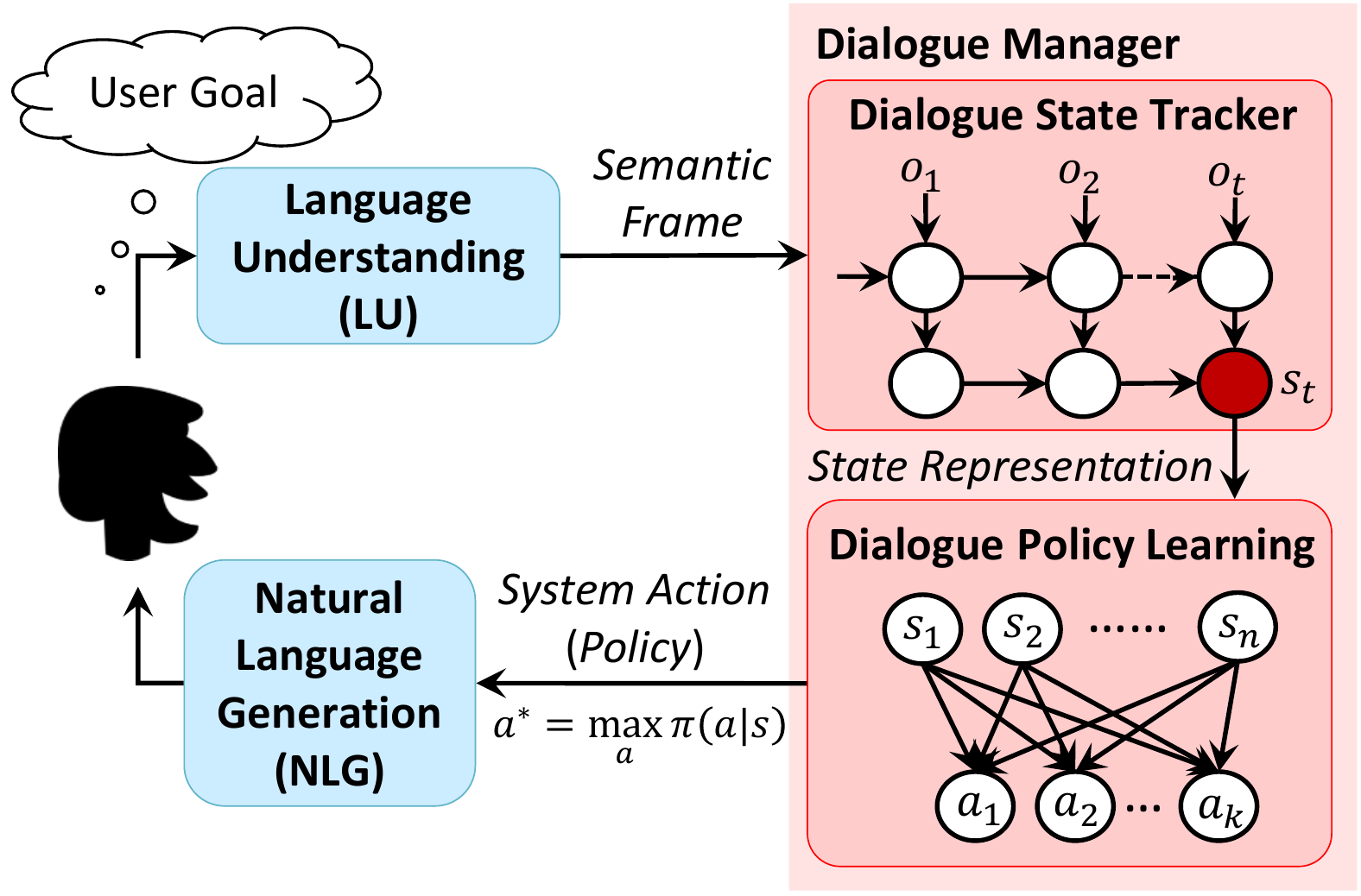}
\vspace{-7mm}
\caption{Illustration of a task-completion dialogue system.}
\label{fig:dialogue_manager}
\vspace{-5mm}
\end{figure}

Figure~\ref{fig:dialogue_manager} illustrates a typical task-completion dialogue system that contains three main components: language understanding (LU) that converts natural language to system-readable semantic frames, natural language generation (NLG) that converts system actions to natural language, and a dialogue manager (DM). The dialogue manager controls state tracking and policy learning, where dialogue policy learning can be regarded as a sequential decision process. The system will learn to select the best response action at each step, by maximizing the long-term objective associated with a reward function.

%%%Figure~\ref{fig:dialogue_manager} illustrates a typical neural dialogue system framework. The dialogue manager consists of two modules, 1) a dialogue state tracker for tracking the evolving states during the conversation, and 2) a policy learner that selects the next action based on the current dialogue state.
This paper focuses on dialogue policy learning (the bottom-right part of Figure~\ref{fig:dialogue_manager}), where the input of the policy learner is the dialogue state representation $s$ that consists of the latest user action (e.g., \textsf{request\_moviename(genre=action, date=today)}), the last agent action (e.g. \textsf{request\_location}), history dialogue turns, and available database results. The learned dialogue policy then helps the agent decide what action $a$ to take in each turn of the conversation, in order to maximize the future cumulative reward. 

As aforementioned, dialogue policy optimization can be formulated as a sequential decision problem to maximize the long term objective associated with a reward function. The advantage actor-critic (A2C) method has achieved superior performance on solving sequential decision problems \cite{mnih2015human,silver2016mastering,silver2017mastering}. Su et al. applied the actor-critic model to dialogue policy optimization and proved its superiority on convergence to other methods such as deep Q-networks~\cite{su2015reward}. Similarly, we employ an actor-critic approach to learn dialogue policy in our model. In addition, inspired by GAN~\cite{goodfellow2014generative} (using a discriminator to guide the training of generative models), we form a minimax game between a generator (an actor that selects actions in our scenario) and a discriminator, to judge whether an action is performed by the expert or the actor. The discriminator can be regarded as another critic and servers as a heuristic intrinsic reward function to guide the actor towards expert-like regions. Another related topic is inverse reinforcement learning~\cite{abbeel2004apprenticeship}, which is to recover the reward function from expert demonstrations, samples of the trajectories executed by experts~\cite{ng2000algorithms}. Ho and Ermon also drew a connection between inverse reinforcement learning and generative adversarial networks to learn the reward function in the GAN framework~\cite{NIPS2016_6391}.
Compared to their work that focused on learning the extrinsic reward, in this paper, we use intrinsic reward to speech up the training.

\subsection{Advantage Actor-Critic for Dialogue Policy Learning}
The training objective of policy-based approaches is to find a policy $\pi$ that maximizes the expected reward $R$ (minimizes the loss $J$) over all possible dialogue trajectories. The expected reward is defined as $R = \sum_{t=0}^{T-1} \gamma^t r_t$ over a dialogue with the length $T$, where $r_t$ is the reward at time stamp $t$, and $\gamma$ is the discount factor. 
The policy $\pi$ is a parametrized probabilistic mapping function between the state space and the action space:
\begin{eqnarray}
\pi_\theta(a\mid s) = P(A_t = a\mid S_t = s; \theta),
\end{eqnarray}
where $\theta$ represents the parameters learned by policy gradient algorithms~\cite{DBLP:journals/ml/Williams92}.
Given the objective function, the gradients of the parameters are computed as
\begin{eqnarray}
\label{eqa:pg}
\nabla_\theta J(\theta) = \mathbb{E} [ \nabla_\theta \log  \pi_\theta(a\mid s) Q^{\pi_\theta}(s, a) ]
\end{eqnarray}
where $Q^{\pi_\theta}(s,a)$ is the long-term reward value.
However, the gradients usually have high variance, which makes the learning task more challenging.
A baseline function $B(s)$ is usually employed to reduce the variance, while keeping the estimated gradient unchanged~\cite{mnih2016asynchronous}.
Here we can simply choose the state value function as a baseline $B(s) = V^{\pi_\theta}(s)$. With this strategy, we can rewrite (\ref{eqa:pg}) using the advantage function $A^{\pi_\theta}(s,a)$:
\begin{eqnarray}
\nabla_\theta J(\theta) &=& \mathbb{E} [\nabla_\theta \log  \pi_\theta(a\mid s) A^{\pi_\theta}(s,a)], \\
A^{\pi_\theta}(s,a) &=& Q^{\pi_\theta}(s,a) - V^{\pi_\theta}(s).
\end{eqnarray}
However, in this setting, there are two functions and parameters that need to be learned.
In order to reduce the number of required parameters and improve stability, temporal difference (TD) error is employed as an unbiased estimate of the advantage function,
\begin{eqnarray}
\label{eqa:a2cvalue}
\delta^{\pi_\theta} = r + \gamma V^{\pi_\theta}(s') - V^{\pi_\theta}(s),
\end{eqnarray}
In this way, the policy gradient with the TD error can be computed as
\begin{eqnarray}
\label{eqa:a2c}
\nabla_\theta J(\theta) &=& \mathbb{E} [\nabla_\theta \log  \pi_\theta(a\mid s) \delta^{\pi_\theta}].
\end{eqnarray}
The policy network $\pi_\theta$ is termed as the \emph{actor} to yield a dialogue system action, and the advantage function $A^{\pi_\theta}$ is the \emph{critic}, indicating ``good'' or ``bad'' for executing an action given a state. 
The classic A2C architecture is shown at the bottom part of Figure~\ref{fig:a2c} without discriminator.

\subsection{Adversarial Model for Dialogue Policy Learning}
\label{sec:adversarial}

\begin{figure}[t]
\centering
\includegraphics[width=1\linewidth]{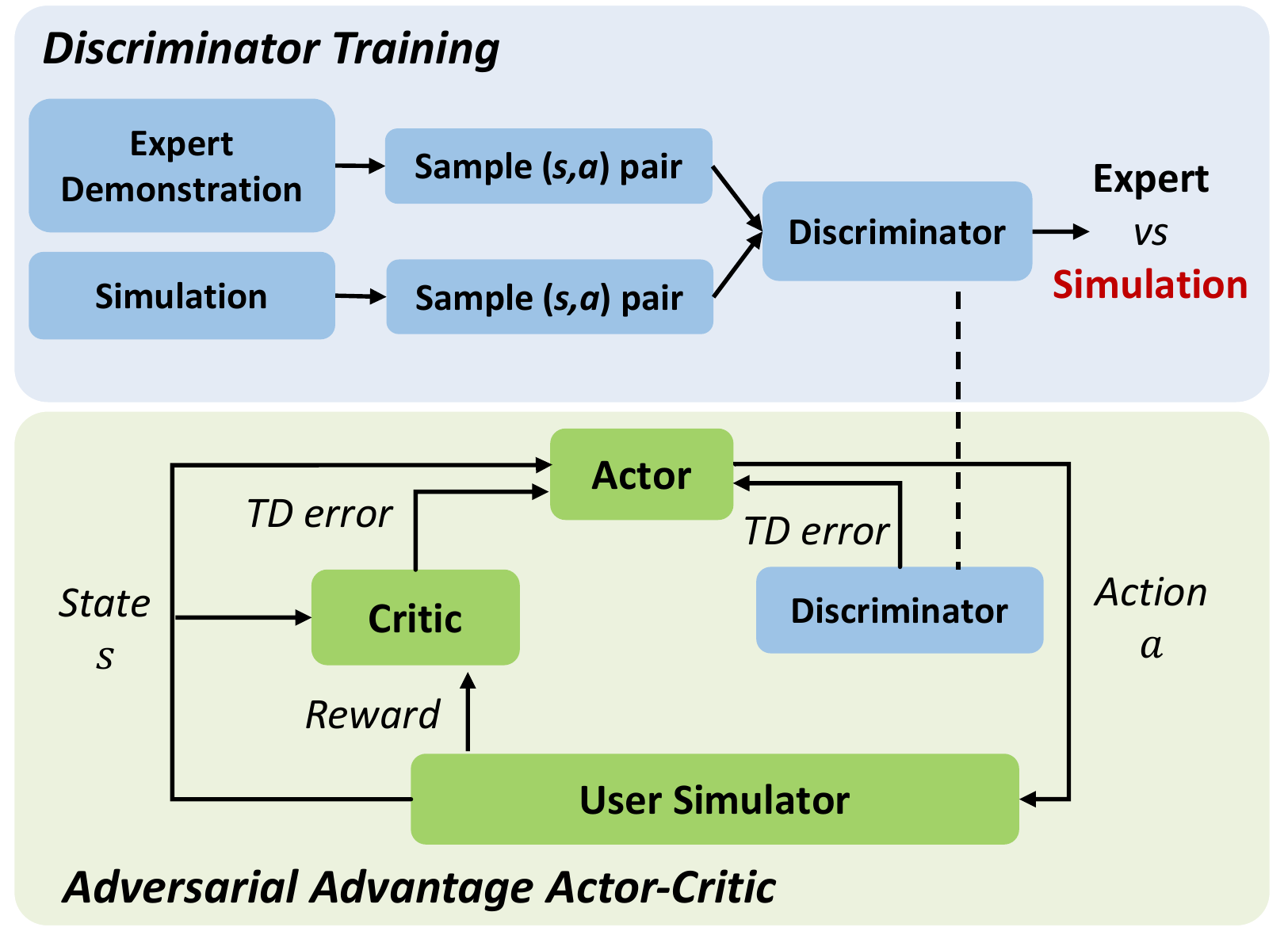}
\vspace{-5mm}
\caption{Illustration of the proposed adversarial advantage actor-critic for dialogue policy learning.}
\label{fig:a2c}
\vspace{-2mm}
\end{figure}

%In task-completion dialogues, the reward function of actor $\pi$ is usually defined as a hand-crafted indicator of success or failure. However, the reward function for dialogue optimization is complex, and the manually defined function may not provide sufficient evidence for efficient learning. Inverse reinforcement learning was applied to recover the reward function from expert demonstrations, which are samples of the trajectories executed by experts~\cite{ng2000algorithms}. Ho and Ermon also drew a connection between inverse reinforcement learning and generative adversarial networks~\cite{NIPS2016_6391}.
GAN is a minimax competing game between a generator and a discriminator.
In our scenario, the actor $\pi$ can be viewed as a generator $G$, which aims to generate actions that can purposefully confuse a discriminator $D$. The discriminator $D$ is expected to identify a state-action pair $(s,a)$ as either an expert demonstration or a simulation experience. When $D$ cannot distinguish actions generated from the actor $\pi$ and those from the experts, we believe that $\pi$ has been improved from the previous state. Moreover, $D$ can be viewed as a reward function extracted from the experts' trajectories. Figure~\ref{fig:a2c} shows the discriminator training procedure using adversarial learning.

The training objective is to find a saddle point $(\pi, D)$ of
\begin{eqnarray}
\mathbb{E}_{\pi} [\log (D(s,a))] + \mathbb{E}_\text{Demo}[\log (1-D(s,a)].
\end{eqnarray}
More specifically, let $\theta_D$ denote the parameters of the discriminator $D$.
The training objective of $D$ is simply to maximize the probability of classifying each state-action pair $(s,a)$ correctly:
\begin{eqnarray}
\label{eqa:dis}
 \min_{\theta_D} \mathcal{L}_D &=& -\mathbb{E}_{(s,a)\sim Simu} \log D(s,a;\theta_D)  \\
&& - \mathbb{E}_{(s,a)\sim Demo} \log (1 - D(s,a;\theta_D))\nonumber 
\end{eqnarray}
where $Simu$ and $Demo$ represent simulation experience and expert demonstration, respectively.
As thus, the actor $\pi$ can be improved using actor-critic, with $-\log (1-D(s,a))$ as the reward function.
The updated gradients can then be reformed as:
\begin{eqnarray}
\label{eqa:a2c_adv}
\nabla_\theta J(\theta) &=& \mathbb{E} [\nabla_\theta \log  \pi_\theta(a\mid s) A^{\pi_\theta}_\text{GAN}(s,a)].
\end{eqnarray}
Similarly, we use TD error as an unbiased estimation of the advantage function:
\begin{eqnarray}
\label{eqa:a2cadvvalue}
\delta^{\pi_\theta}_\text{GAN} = r_\text{GAN} + \gamma V^{\pi_\theta}_\text{GAN}(s') - V^{\pi_\theta}_\text{GAN}(s).
\end{eqnarray}

\subsection{Adversarial Advantage Actor-Critic}

Furthermore, training dialogue policy with a stand-alone adversarial model can be impractical, due to the high dimensionality of its state-and-action space. 
To address this issue, we propose the adversarial advantage actor-critic (Adversarial A2C) method as depicted in Figure~\ref{fig:a2c}, which combines A2C with a reward function learned from an adversarial model that serves as another additional critic for the actor $\pi$. 
There are several ways to combine two critics, such as linear combination of two reward functions or alternately optimizing with each reward function. 
In our experiments, we use alternating optimization.
Algorithm~\ref{algo:a3c} outlines the full procedure of training the Adversarial A2C model.
The goal is to encourage the actor to select better actions guided by a discriminator, in order to improve the efficiency and effectiveness of the exploration.

%something like this - the reward function parameterized by $\theta_r$, GAN model updates the reward parameters by estimating the gradient of the objective return with the reward function from experts' and simulation experience. 

\begin{algorithm}[t]
\small
\caption{Adversarial Advantage Actor-Critic Model} % for Dialogue Policy Learning}
\begin{algorithmic}[1]
\STATE \textbf{Input: } Expert demonstrations $Demo$, initialize actor $\pi$, discriminator $D$ and two value functions $V^{\pi_\theta}(s)$, $V^{\pi_\theta}_\text{GAN}(s)$
\FOR {$i$=$1$:$N$}
\STATE Restart the dialogue simulator, get state representation $s$, initialize transition tuple \emph{buffer} = []
\WHILE{$s$ is not a terminal state}
\STATE Perform the action $a_t$ according to the actor $\pi(a_t\mid s;\theta)$
\STATE Receive the reward $r_t$ and switch to a new state $s'$
\STATE Store $(s, a_t, r_t, s')$ to the transition tuple \emph{buffer}
\STATE $s := s'$
\ENDWHILE
\STATE Train the actor $\pi$ with gradients (\ref{eqa:a2c})
\STATE Train value function $V^{\pi_\theta}(s)$ by minimizing the TD error (\ref{eqa:a2cvalue})
\STATE Sample state action pairs $(s,a)$ from expert demonstration $Demo$
\STATE Train the actor $\pi$ with gradients (\ref{eqa:a2c_adv})
\STATE Update the reward with $-\log (1-D(s,a))$ and train value function $V^{\pi_\theta}_\text{GAN}(s)$ by minimizing the TD error (\ref{eqa:a2cadvvalue})
\STATE Update the discriminator parameters (\ref{eqa:dis})
\ENDFOR
\end{algorithmic}
\label{algo:a3c}
\vspace{-1mm}
\end{algorithm}

\section{Experiments}
\label{sec:exps}
To verify the performance of the proposed model, we evaluated it in a task-completion dialogue system for movie-ticket booking. In this system, the agent will gather information from users through conversations and eventually book the movie tickets for them. The environment then judges a binary outcome (success or failure) at the end of each conversation, based on: 1) whether a movie ticket is booked, and 2) whether the booked ticket satisfies the constraints requested by the user.

\subsection{Experimental Setup}
%\subsection{Dataset}
The dataset used in our experiment is raw conversational data collected via Amazon Mechanical Turk, annotated by domain experts~\cite{li2017end}.
This single-domain movie-ticket booking dataset contains 11 dialogue acts and 29 slots, including \emph{informable} slots (users can use these to narrow down the search), and \emph{requestable} slots (where users can ask the agent for more information).
There are in total 280 labeled dialogues, with an average length of 11 turns.

%\subsection{User Simulator}
In order to perform end-to-end training for the dialogue system, a user simulator is required to interact with the system in a natural way.
We adopted a publicly available, user-agenda-based simulator in our experiments~\cite{li2016user}.
In a task-completion dialogue setting, the user simulator first generates a user goal, and the dialogue agent tries to help the user accomplish that goal in the course of the conversation, without explicitly knowing the user goal.
A user goal normally consists of two parts: 
\emph{inform\_slots} representing slot-value pairs that serve as constraints from the user, and
\emph{request\_slots} representing slots whose value the user has no information about, but wants to get information from the agent through the conversation. In our experiment, the user goals were generated from labeled conversational data.

%\paragraph{User Agenda Modeling:}
%\label{sec:user_agenda}
%\textbf{User Agenda Modeling:} During the course of a dialogue, the user simulator maintains a compact, stack-like representation called \emph{user agenda}~\cite{schatzmann2009hidden}, where the user state $s_u$ is factored into an agenda $A$ and a goal $G$. The goal consists of constraints $C$ and request $R$. At each time-step $t$, the user simulator generates the next user action $a_{u,t}$ based on the current state $s_{u,t}$ and the last agent action $a_{m,t-1}$, and then updates the current status $s'_{u,t}$.

\subsection{Implementation}
In Figure~\ref{fig:a2c}, the expert demonstrations can be collected from either human or pre-trained agent. In our experiment, we collected 50 successful dialogues from a pre-trained agent. The discriminator is a binary classifier of a single-layer neural network with 80 hidden units. For the actor, we use a single-layer neural network with a hidden size of 80, pre-trained with rule-based examples in order to give acceptable initialization. During the Adversarial A2C model training, two critics (the critic and the discriminator in Figure~\ref{fig:a2c}) are applied alternatively, where their value functions are 
%with the predefined reward and the value function with reward function learned from GAN are 
single-layer neural networks with 80 hidden units. All parameters are optimized with \emph{RMSProp}. During training, the model is updated at the end of each dialogue episode.
% it is actually value function, not q value function.
%state transition tuples are accumulated and the model is updated at the end of each dialogue episode.

%For the actor, we use a single-layer neural network with hidden size 80, it is pre-trained with some rule-based examples to give a good initial start. The discriminator is a binary classifier with 80 hidden units, single-layer neural network. The two critics (Critic and Discriminator in the Figure~\ref{fig:a2c}) are applied alternatively in the training. The value function with predefined reward and value function with reward learned for GAN are single-layer neural networks with 80 hidden units. All the parameters are optimized with \emph{RMSProp}. State transition tuples are accumulated and the model is updated at the end of each dialogue episode. The demonstrations from experts can be either collected from human or a pre-trained agent. In this paper, we collect 50 successful dialogues from an agent pre-trained off-line.

%Training the actor from scratch on-line will always obtain poor performance, which is a cold start issue.

\subsection{Evaluation Results}
In the movie-ticket booking task, we benchmark the proposed Adversarial A2C model against three baseline models on three metrics: success rate, average rewards, and the average number of turns per dialogue session.
\begin{compactitem}
\item \emph{Rule Agent} is a handcrafted rule-based policy that informs and requests a hand-picked subset of necessary slots.
\item \emph{A2C Agent} is trained with a pre-defined reward function and a standard advantage actor-critic algorithm.
\item \emph{BBQN-Map Agent} is the best agent among a set of BBQN variants (including BBQN-VIME) and DQN variants, which has demonstrated great efficiency for policy exploration in task-completion dialogue systems~\cite{lipton2016efficient}.
\vspace{2mm}
\end{compactitem}

\begin{figure}[t]
\centering
\vspace{-5mm}
\includegraphics[width=1.05\linewidth]{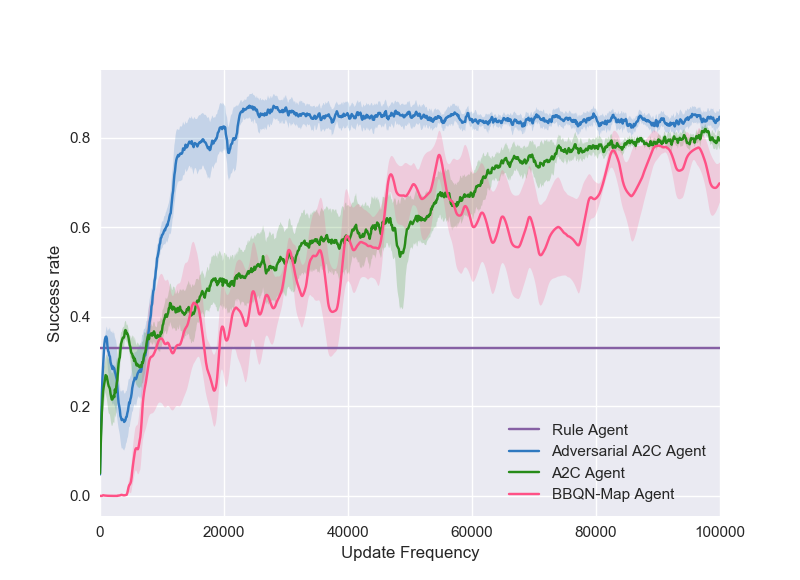}
\vspace{-8mm}
\caption{Learning curves of dialogue policies.}
\label{fig:learning_curve}
\vspace{-1mm}
\end{figure}

%\begin{figure}[htb]
%\centering
%\includegraphics[width=1\linewidth]
%{figures/learning_curve_reward.png}
%\vspace{-3mm}
%\caption{Learning curves of dialogue policies, each learning curve is averaged over 5 runs.}
%\label{fig:learning_curve}
% %\vspace{-3mm}
%\end{figure}
 
\begin{table}[t]
%\small
\centering
\begin{tabular}{lccc}
\toprule
{Agents} & { Success Rate} & { Reward} & {Turn} \\ 
\midrule
{Rule} & 41.34 & 0.26 & 16.00\\
{A2C} & 81.24 & 5.08 & 15.43  \\
{BBQN-Map} & 81.56 & 5.00 & 18.75  \\
{Adversarial A2C} & \textbf{87.52} & \textbf{5.93} & \textbf{13.52} \\
\bottomrule
\end{tabular}
\vspace{-1mm}
\caption{Final agent performance on 5K simulated dialogues.}
\vspace{-2mm}
\label{tab:final_results}
\end{table}

Figure~\ref{fig:learning_curve} shows the learning curves of all these dialogue agents mentioned above, and Table~\ref{tab:final_results} shows the evaluation performance of each agent, averaged over 5 runs.
%Table~\ref{tab:final_results} shows the testing performance averaged over 5 runs.
The learning curves in Figure~\ref{fig:learning_curve} shows that Adversarial A2C agent can learn much faster with better exploration capability. The learning curve is also more stable compared with others. Table~\ref{tab:final_results} suggests that the Adversarial A2C agent can yield better dialogue policies than other approaches, in terms of success rate, average reward, and average number of turns per dialogue. 

\vspace{-1mm}
\section{Conclusions}
\label{sec:conclusion}
This paper presents an adversarial advantage actor-critic model, which can explore policy learning in task-completion dialogue systems with great efficiency. The proposed model learns a discriminator from expert demonstrations and online experience, and then the learned discriminator serves as an additional critic to guide policy learning. Our experiments in a movie-ticket booking domain demonstrate the superiority and efficiency of the proposed model in policy learning, compared with state-of-the-art approaches. The promising results suggest several interesting future directions: 1) employing variance-reducing methods to stabilize the gradient calculation, in order to address the high variance issue in policy gradient estimation, 2) applying the model to more complicated dialogue tasks, such as composite task-completion dialogues~\cite{peng2017composite}, and 3) extending this work to other deep reinforcement learning benchmark tasks and other domains.

\vfill\pagebreak

\small
\bibliographystyle{IEEEbib}
\bibliography{refs}

\end{document}